%% file: acl_latex.tex
\documentclass[11pt]{article}

\usepackage[preprint]{acl}
\usepackage{times}
\usepackage{latexsym}

\usepackage[T1]{fontenc}

\usepackage[utf8]{inputenc}

\usepackage{microtype}

\usepackage{inconsolata}

\usepackage{graphicx}

\usepackage{amssymb}
\usepackage{mathtools}
\usepackage{amsthm}
\usepackage{enumitem}
\usepackage{wrapfig}
\usepackage{url}
\usepackage{microtype}      
\usepackage{dsfont}
\usepackage[framemethod=TikZ]{mdframed} 
\usepackage{multirow}
\usepackage{multicol}
\usepackage{graphicx}
\usepackage{longtable}
\usepackage{booktabs} 
\usepackage{makecell}
\usepackage{mathabx}
\usepackage{enumitem}
\usepackage{listings}
\usepackage{pifont}
\usepackage{tikz}
\usepackage{pgfplots}
\usepackage[export]{adjustbox}
\usepackage{caption}
\usepackage{subcaption}
\usepackage{tabularx}
\usepackage{fontawesome}

\usepackage{colortbl}
\usepackage{tabulary}
\usepackage{etoolbox}
\definecolor{palegreen}{rgb}{0.40, 0.68, 0.40}
\definecolor{aliceblue}{rgb}{0.94, 0.97, 1.00}
\definecolor{honeydew}{rgb}{0.94, 1.00, 0.94}

\definecolor{promptcolor}{RGB}{200, 235, 227}
\definecolor{prompttitlecolor}{RGB}{175, 172, 172}

\usepackage[normalem]{ulem}
\usepackage{tcolorbox}
\usepackage{listings}
\tcbuselibrary{skins,breakable}
\usepackage{tikz}\usetikzlibrary{positioning,arrows.meta,fit}
\DeclareCaptionType{listing}[Figure][List of Listings]
\mdfdefinestyle{prompt}{%
    linecolor =   black,
    linewidth =   0.2pt,
    backgroundcolor =  promptcolor,
    roundcorner     =   1pt,
    font = \small\ttfamily,
    innerleftmargin=1.5pt,
    innerrightmargin=1.5pt
}

\newmdenv[%
    roundcorner=5pt, 
    linecolor =   black,
    linewidth =   1pt,
    font = \small\ttfamily,
    subtitlebackgroundcolor=prompttitlecolor, 
    frametitlebackgroundcolor=prompttitlecolor,
    backgroundcolor=promptcolor, 
    frametitle={Generated caption},
    subtitleaboveskip=0.5\baselineskip,
    subtitlebelowskip=0.5\baselineskip,
    ]{captionenv}

\DeclareMathAlphabet{\pazocal}{OMS}{zplm}{m}{n}

\DeclareMathAlphabet\mathbfcal{OMS}{cmsy}{b}{n}

\definecolor{topcolor}{rgb}{1,0.8,0.8}
\definecolor{secondcolor}{rgb}{1,0.87,0.7}
\definecolor{thirdcolor}{rgb}{1,1,0.8}
\definecolor{promptcolor}{RGB}{200, 235, 227}
\definecolor{prompttitlecolor}{RGB}{175, 172, 172}

\author{Tao Zhang\\
Microsoft Research Asia\\
\texttt{zhangt@microsoft.com}
\And
Yuhang He\thanks{Corresponding author.}\\
Microsoft Research Asia\\
\texttt{yuhanghe@microsoft.com}
}

\title{LongCapHarness: Capability-Guided Auto-Harnessing \\for Long-Context Reasoning}

\begin{document}
\maketitle

\input{secs/sec0_abstract}
\input{secs/sec1_intro}

\input{secs/sec2_relatedwork}

\input{secs/sec4_method}

\input{secs/sec5_exp}
\input{secs/sec5.1_limit}

\bibliography{main}

\appendix

\input{secs/sec6_appendix}

\end{document}

%% file: secs/sec0_abstract.tex
\begin{abstract}

Existing methods enhancing an LLM's long-context reasoning capability often rely on data synthesis and follow-up LLM post-training. However, the data synthesis strategy and the post-training recipes are empirically driven and computationally intensive. In this work, we propose a new framework, dubbed \emph{LongCapHarness}, to instead harness the underlying capabilities for long-context reasoning. The harnessing process recursively revises and updates capability primitive implementation, capability state and control logic through recursive self-improvement on the development set. With neither data synthesis nor LLM post-training, \emph{LongCapHarness} is applied to Llama3-8B and transferred to Qwen2.5-3B. \emph{LongCapHarness} substantially outperforms recent data-synthesis-based baselines on HELMET and RULER benchmarks; its estimated development cost is less than 50\% of the estimated training costs of NExtLong and PolicyLong. We hope these results motivate further research into harnessing the latent capabilities of LLMs to solve tasks beyond their explicit training.
\end{abstract}

%% file: secs/sec1_intro.tex
\section{Introduction}
\label{sec:introduction}

Existing methods for enhancing a pretrained LLMs' long-context reasoning capabilities follow the data synthesis pipeline: synthesizing a large long-context corpus, followed by post-training the LLM on the synthesized data~(\textit{e.g.}, NExtLong~\cite{gao2025nextlong}, PolicyLong~\cite{jia2026policylong}, LongFilter~\cite{deng2026beyond}, EntropyLong~\cite{jia2026entropylong}, LADM~\cite{chen2025ladm}, Quest~\cite{gao2025quest}, LiteLong~\cite{jia2026litelong}). Despite the observed gains on long-context benchmarks such as HELMET~\cite{yen2025helmet} and LongBench-v2~\cite{bai2025longbenchv2}, these methods disentangle data synthesis from the follow-up LLM post-training, and the data synthesis strategies and post-training recipes are empirically motivated and computationally intensive. Therefore, the direction on data synthesis and LLM post-training is costly and difficult to scale-up.

\begin{figure}[t]
    \centering
    \includegraphics[width=0.99\linewidth]{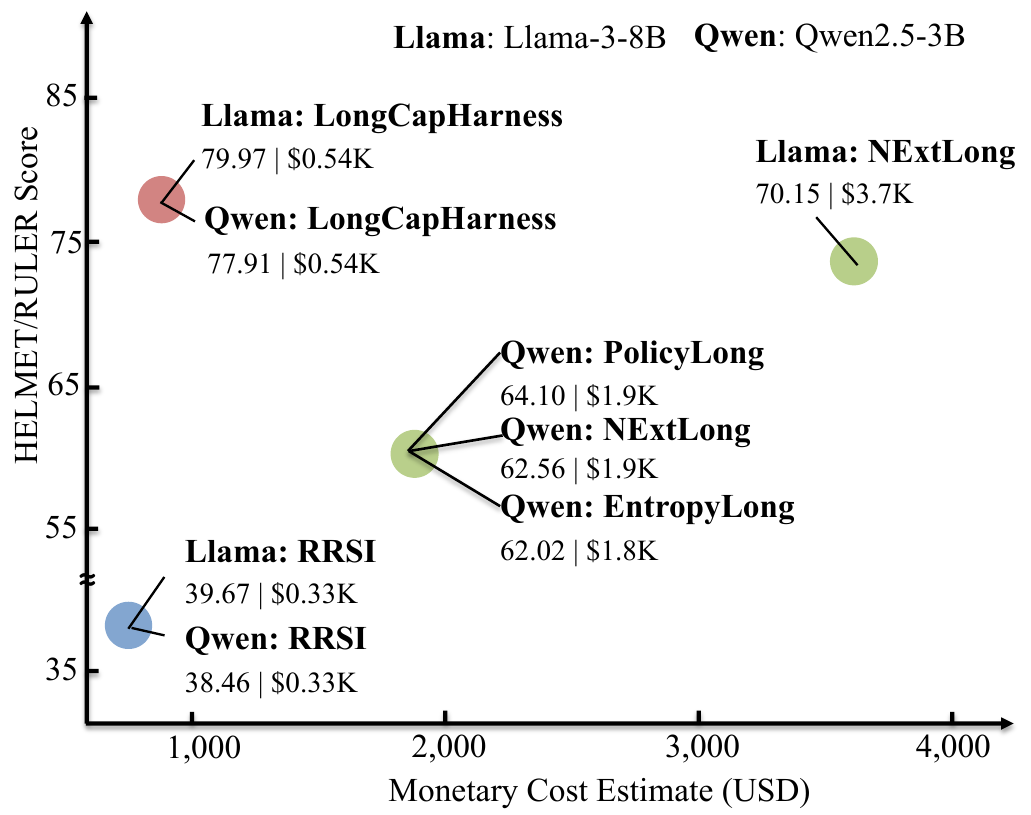}
    \caption{Performance superiority and cost effectiveness of \emph{LongCapHarness} compared with the three recent data-synthesis-based baselines and the RSI baseline.}
    \label{fig:bubble_compare}
    \vspace{-5pt}
\end{figure}

In this work, we propose an alternating framework, dubbed \emph{LongCapHarness}, that substantially outperforms data-synthesis-based methods while incurring much lower development cost~(see Fig.~\ref{fig:bubble_compare}). \emph{LongCapHarness} builds on the key observation:


\begin{tcolorbox}[colframe=blue!40, colback=gray!5,boxrule=0.3mm, fontupper=\textit,]
\emph{LLMs without prior explicit long-context pre-training can still show strong performance on long-context benchmarks as long as they have been appropriately equipped with long-context reasoning capability.}
\label{box:longcapharness_insight}
\end{tcolorbox}

To build up such a framework, we leverage the latest harness engineering technologies~\cite{chen2026harnessx,lee2026metaharness,lin2026ahe,lou2026autoharness,nie2026tthe} to design a long-context capability-guided auto-harnessing framework. We argue that the central challenge in long-context reasoning is not merely fitting a long input within the available context window. Solving such a task often requires identifying its task structure, localizing relevant evidence, selectively retaining intermediate results, and finally integrating them into a globally consistent answer. In a monolithic end-to-end call, these operations and dependencies remain implicit, making it hard for the model to coordinate them while processing the full input at once. To address this challenge, we propose a capability harness to support long-context tasks. Capability decomposition further makes this latent task structure explicit by breaking the long-context task into operations with defined inputs and outputs. The harness can then preserve shared state across operations and supply each model call only with the context it needs, decoupling total input length from the context budget of any individual call.

In \emph{LongCapHarness}, we leverage a harnessing agent to construct reusable capability harnesses from the development set.
The harness agent studies the task input to identify the required capability, then decomposes it into capability primitives. The harnessing process recursively revises and updates the primitive implementation, capability state and control logic by comparing the answer generated by running the capability harness against the reference answer. The final capability harness is frozen and saved to a library. At test time, an execution LLM selects and runs the appropriate capability harness on the held-out test set.


We evaluate \emph{LongCapHarness} on HELMET~\citep{yen2025helmet}, RULER~\citep{hsieh2024ruler}, and LongBench-v2~\citep{bai2025longbenchv2}, using GPT-5.6-Sol~\citep{openai2026solpricing} for harness development and Llama3-8B~\citep{meta2024llama3} for execution before transferring the harness to Qwen2.5-3B~\citep{yang2024qwen25}. On HELMET and RULER, it outperforms the corresponding model-training baselines~\citep{gao2025nextlong,jia2026policylong} and RRSI~\citep{xia2026rrsi} with both execution models. On LongBench-v2, Qwen2.5-3B with the harness surpasses PolicyLong~\citep{jia2026policylong} on the Medium subset but trails it on the Long subset.

Estimated development API costs are lower than the estimated training costs of NExtLong and PolicyLong, but higher than RRSI development costs. At inference, \emph{LongCapHarness} uses substantially fewer output tokens than RRSI; estimated costs are comparable with Llama but higher with Qwen, where increased input-token use offsets the output savings.

Our contributions are summarized as follows:
\begin{enumerate}
\item We propose \emph{LongCapHarness}, a novel framework for long-context reasoning tasks with neither long-context data synthesis nor LLM post-training.
\item We demonstrate the superiority of \emph{LongCapHarness} in terms of performance on benchmarks and cost effectiveness via extensive experiments.
\item We hope to motivate more research into harnessing the latent capabilities of LLMs to solve complex tasks that go beyond their explicit training.
\end{enumerate}

%% file: secs/sec2_relatedwork.tex
\section{Related Work}
\label{sec:related-work}


\paragraph{Data-Synthesis-based Long-Context Reasoning.}
Most recent methods on improving long-context reasoning first synthesize long-context data and then use it to post-train LLMs~\cite{jia2026policylong,jia2026entropylong,gao2025nextlong,chen2025ladm,gao2025quest}. These data strategies are often developed empirically and separately from the post-training recipes. This separation makes it difficult to determine which data properties are essential for long-context reasoning and how data preparation should be aligned with post-training. For example, ProLong \cite{gao2025train} studies training recipes and data
mixtures, while NExtLong \citep{gao2025nextlong} constructs long
sequences with hard-negative distractors.
EntropyLong \cite{jia2026entropylong} verifies useful context
through predictive uncertainty, and PolicyLong
\cite{jia2026policylong} refreshes data using the evolving model.
\emph{LongCapHarness} instead proposes an alternating direction with neither data synthesis nor LLM post-training. Benefiting from its carefully designed capability harnessing framewor, \emph{LongCapHarness} exhibits both performance superiority and cost effectiveness compared with these data-synthesis-based methods.

\paragraph{Model--program computation and semantic operators.}
Model--program systems combine structured computation with
learned semantic operations.
SUQL \citep{liu2024suql} extends SQL with free-text question
answering and summarization, while LOTUS \citep{patel2025lotus}
provides optimized semantic operators for filtering, joining,
aggregation, and ranking.
\emph{LongCapHarness} uses task capabilities to guide how computational
inputs are recovered from long documents and composed into
executable workflows.

\paragraph{Agent and harness design.}
Automated agent design improves model-based systems by searching
and refining external execution logic.
ADAS \citep{hu2024adas} searches over code-defined agents using
past designs and evaluations. AFlow \citep{zhang2024aflow}
optimizes workflows through Monte Carlo Tree Search and execution
feedback, and AutoHarness \citep{lou2026autoharness} synthesizes
harnesses and policies from game-environment feedback.
\emph{LongCapHarness} adds explicit capability contracts to guide
the construction and refinement of representations, operators,
and model calls for long-context reasoning.

%% file: secs/sec4_method.tex
\section{\emph{LongCapHarness} Introduction}
\label{sec:methodology}

\begin{figure}[h]
    \centering
    \small
    \includegraphics[width=\linewidth]{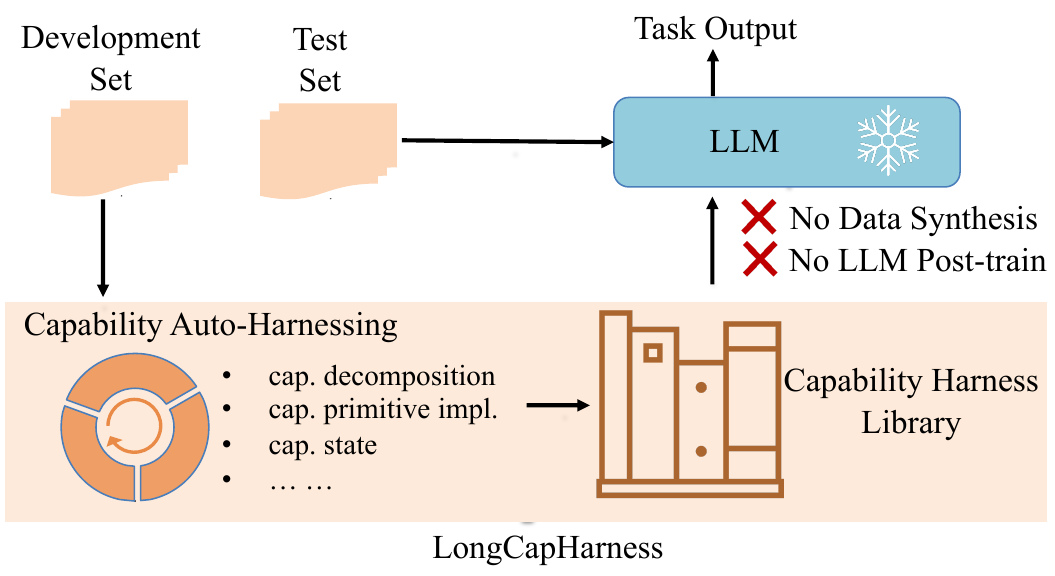}
    \caption{\emph{LongCapHarness} uses recursive self-improvement~(RSI) on the development set to construct and refine reusable capability harnesses. The resulting library is used at test time to assist an LLM execution on long-context tasks. \emph{LongCapHarness} requires neither data synthesis nor LLM post-training.}
    \label{fig:method-overview}
\end{figure}

\emph{LongCapHarness} starts with a development set of task inputs and corresponding reference answers. The harnessing agent~(in our case, we adopt GPT-5.6-Sol~\citep{openai2026gpt56sol}) studies the task, identifies the required capability, and further decomposes it into capability primitives~(Sec.~\ref{sec:capdefine}). The harnessing agent recursively revises and updates the primitive implementation, capability state and control logic in recursive self-improvement~(RSI)~(Sec.~\ref{sec:harnes_process}). After the harnessing procedure, we freeze all capability harnesses and save them to the capability harness library. Each capability harness in the library is associated with a brief text description showing the capability summary, input and output information. During test-time task execution, the capability harness library is selectively loaded and reused to complete the task in the test set~(Sec.~\ref{sec:test-time-execute}). The whole capability auto-harnessing pipeline is shown in Fig.~\ref{fig:method-overview}.

\subsection{Capability in \emph{LongCapHarness}}
\label{sec:capdefine}

A capability $\mathcal{C}$ in \emph{LongCapHarness} is defined as an essential \textit{abstract transformation} from the an input $\mathcal{I}$ to an output $\mathcal{O}$,

\begin{equation}
    \mathcal{O} = \mathcal{C}(\mathcal{I}),\quad \mathcal{I} \neq \mathcal{O}
\label{eqn:capdefine}
\end{equation}

\begin{figure}[t]
    \centering
    \includegraphics[width=1.0\linewidth]{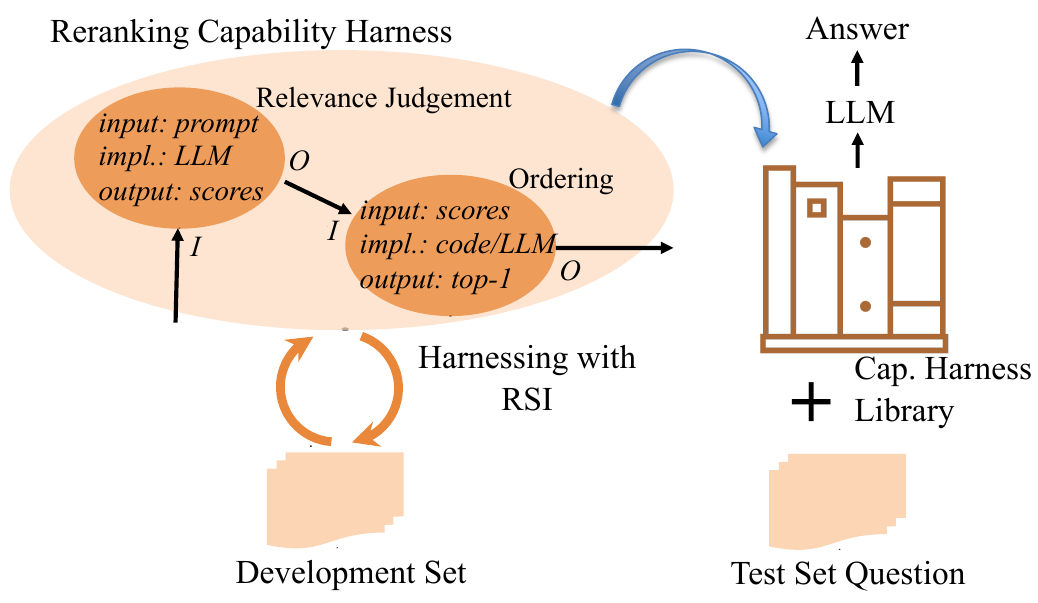}
    \caption{Capability Illustration: we use the \textit{Reranking Capability} as an example. It is divided into \textit{Relevance Judgement} and \textit{Ordering} capability primitives with sequential control logic. The auto-harnessing on development set with recursive self-improvement~(RSI) implements the two capability primitives, and finally returns the Reranking Capability Harness which is saved in the capability library.}
    \label{fig:cap_harness_illu}
\end{figure}

The capability in Eqn.~(\ref{eqn:capdefine}) abstracts the transformation and is independent of a particular Q\&A pair. Two Q\&A pairs can share the same capability if they require the same transformation. For each capability, \emph{LongCapHarness} constructs its corresponding capability harness: the peripheral concrete, executable workflow that specifies how to perform that transformation. We define the harness as follows,

\begin{equation}
    \mathcal{A} \leftarrow \mathcal{H}_{\mathcal{C}}(\mathcal{Q})
\label{eqn:harnessdefine}
\end{equation}

\noindent where $\mathcal{H}_{\mathcal{C}}$ implements the harness for capability $\mathcal{C}$. It usually includes the code, prompt templates, and execution rules needed to produce the answer. To construct a capability harness, the harnessing agent decomposes the capability $\mathcal{C}$ into multiple capability primitives. Each primitive $\mathcal{C}_{\mathcal{P}_i}$ abstracts a sub-transform and specifies its input $I_i$ and output $O_i$:

\begin{equation}
    \mathcal{C} := \{\mathcal{C}_{\mathcal{P}_1}, \cdots, \mathcal{C}_{\mathcal{P}_m}\};\quad
    \mathcal{O}_i \leftarrow \mathcal{C}_{\mathcal{P}_i}(I_i)
\label{eqn:primitive}
\end{equation}

\noindent where $i\in \mathbb{Z}^{+}$ and $i\leq m$. The input to a capability primitive may include the original question, selected task records, or outputs from earlier primitives; its output may be an intermediate result or part of the final answer. The capability harnessing in Eqn.~(\ref{eqn:harnessdefine}) then becomes capability primitive harnessing and the control logic $\mathbfcal{F}$ harnessing,

\begin{equation}
    \mathcal{A} \leftarrow \mathbfcal{F}(\{\mathcal{H}_{\mathcal{C}_{\mathcal{P}_i}}(I_i)\}_{i=1}^m)
\end{equation}

We use the \textit{Reranking Capability}, which has been extracted by \emph{LongCapHarness}, to illustrate our capability design in Fig.~\ref{fig:cap_harness_illu}. Based on the development set, the auto-harnessing process identifies the \textit{Reranking Capability} and decomposes it into \textit{Relevance Judgement} primitive and \textit{Ordering} with sequential control logic. The auto-harnessing implements the two capability primitives and recursively improves their harness through recursive self-improvement. The resulting \textit{Reranking Capability Harness} is saved into capability library for potential use on the held-out test set.

\subsection{Harnessing Process in \emph{LongCapHarness}}
\label{sec:harnes_process}

The harnessing agent takes each Q\&A pair in the development set as input. By studying the task description from the input question, the harness agent identifies the task's input and output and decides to propose a new capability~(see Eqn.~(\ref{eqn:capdefine})) or leverage an existing capability to address the question. Once the required capability is decided, the harnessing agent tries to decompose it into capability primitives~(see Eqn.~(\ref{eqn:primitive})) and identify corresponding the control logic governing those primitives. The harnessing agent then implements each capability primitive so that each primitive becomes an executable operation. Executing those capability primitives with the governing control logic provides us with capability harness execution generated answer. We compare it with the reference answer to continue to revise and improve the capability harness. 

In \emph{LongCapHarness}, we adopt recursive self-improvement~(RSI~\cite{xia2026rrsi,chen2026harnessx,lee2026metaharness,lin2026ahe,lou2026autoharness,nie2026tthe}) to continually improve the capability harness. The RSI loop iteratively revises the primitive implementations, their control logic, and the capability state to improve the capability harness.

In \textit{Primitive Implementation}, the harnessing agent implements a capability primitive via either code or an LLM call, depending on the exact primitive as well as its input and output requirement. Code implementation is usually adopted where the primitive execution rules are explicit and without ambiguity, such as assignment, occurrences counting and sorting operations; LLM implementation is suitable when the capability primitive's input and output can be represented by text. The harnessing agent then just requires to specify the text to supply, a prompt template and the response format.

In \textit{Control Logic}, the harnessing agent writes the corresponding control logic to combine the capability primitive implementations into a capability harness. The control logic orchestrates the capability primitive operations execution, manage the input/output in-between and further decides the execution stop condition. The capability primitive operations execution governed by control logic generates the capability harness guided answer for the given input question. Comparing it with the reference answer provides important signal to improve the capability harness.

In \textit{Capability State}, the reusable protocol within a capability is treated as the capability state. It defines how each new input question is converted into a format that is directly usable by capability primitives, what intermediate primitive output needs to be retained to derive the final answer. It is worth noting that capability state is agnostic to the actual input from the development set. The per-input runtime state such as the actual question and intermediate results for one input is created fresh during each run and thus does not count as capability state.

\begin{figure}[t]
    \small
    \centering
    \includegraphics[width=1.0\linewidth]{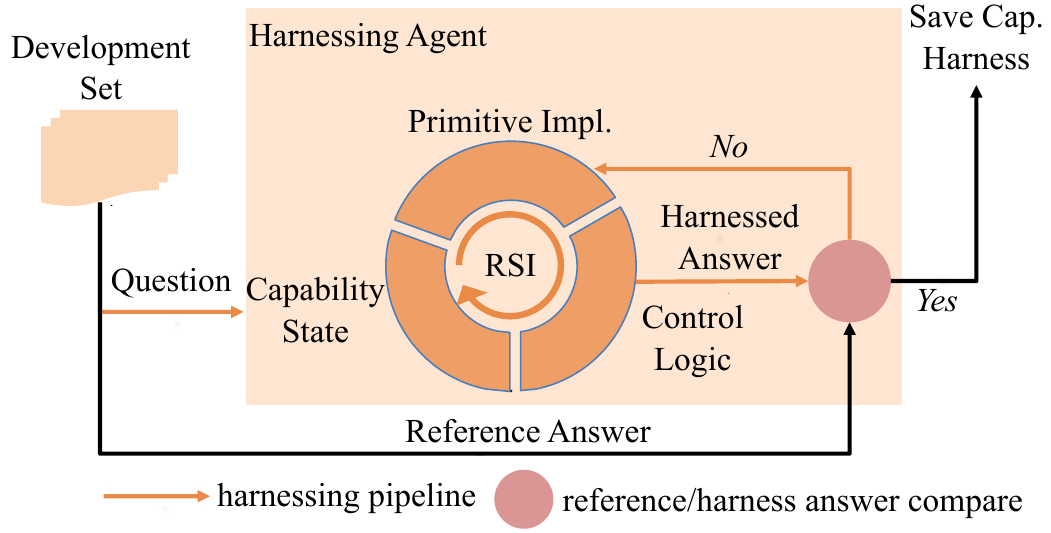}
    \caption{Harnessing Illustration: given the input question, the harnessing agent recursively revise and update the primitive implementation, control logic and capability state until the answer by executing the capability harness is close enough to the reference answer.}
    \label{fig:harness_pipeline_vis}
\end{figure}

In summary, primitive implementations compute results, while control logic schedules and combines those results. The capability state provides the shared workspace that allows these components to exchange outputs without losing the source identity. As shown in Fig.~\ref{fig:harness_pipeline_vis}, during the capability harnessing with RSI, the harnessing agent recursively revises and updates primitive implementation, control logic and capability state until the difference between the answer generated by running the capability and the reference answer is small enough~(the stop condition is decided by the harnessing agent). The resulting capability harness, including primitive implementation, control logic and capability state, is frozen and saved to the capability library. Each capability harness is associated with a text description summarizing the capability, input and output information, this text description is used during the test-time task execution to select the appropriate capability harness to execute the task.

\subsection{Test-Time Task Execution}
\label{sec:test-time-execute}

With the capability harness library obtained in Sec.~\ref{sec:harnes_process}, we execute the task in the held-out test set. Given the question from one Q\&A pair in the test set, the task execution LLM~(we adopt Llama3-8B~\citep{meta2024llama3} and Qwen2.5-3B~\citep{yang2024qwen25}) first selects the capability harness entry by comparing the capability harness text description with the question. 
After selection, the runtime loads the selected harness's code, prompt templates, and execution rules, and passes the question together with its source text and task information as input to the workflow. The harness is executed as a program, where its control logic initializes fresh per-input state and schedules the capability primitives. For each LLM-based primitive, the harness constructs a bounded prompt from that primitive's template, the question or relevant question fields, and the passages or intermediate results needed for the operation. Code-based primitives operate directly on the stored records and intermediate results. The harness retains these results outside the LLM context and supplies them to subsequent primitives until the workflow produces the final answer.

%% file: secs/sec5_exp.tex
\section{Experiments}
\label{sec:experiments}

\begin{table*}[t]
    \centering
    \footnotesize
    \setlength{\tabcolsep}{2pt}
    \caption{\textbf{Main results on HELMET and RULER.}
    We compare LongCapHarness, RRSI, and model-training methods
    with Llama3-8B and Qwen2.5-3B.
    Bold indicates the
    highest score among all methods.
    Dashes indicate that no harness was used or that results were unavailable.}
    \label{tab:main-results}
    \resizebox{\linewidth}{!}{%
    \begin{tabular}{@{}llc|r|*{7}{r}@{}}
        \hline
        Model & Harness & Context & Avg & H-Avg & Recall & RAG & ICL &
        Rank & LongQA & RULER \\
        \hline
        \multicolumn{11}{l}{\textit{Zero-shot}} \\
        Llama3-8B & -- & 8K & 15.14 & 12.48 &
        18.00 & 12.68 & 13.60 & 7.74 & 10.38 & 17.80 \\
        Qwen2.5-3B & -- & 32K & 61.91 & 49.01 &
        -- & -- & -- & -- & -- & 74.81 \\
        \hline
        \multicolumn{11}{l}{\textit{Data Synthesis + LLM Post-training}} \\
        Llama3-8B-NExtLong-128K & -- & 128K & 70.15 & 58.80 &
        82.56 & 60.91 & 81.76 & 31.47 & \textbf{37.30} & 81.50 \\
        Llama3-8B-EntropyLong-128K & -- & 128K & -- & -- &
        -- & -- & -- & -- & -- & 87.37 \\
        Qwen2.5-3B-NExtLong-128K & -- & 128K & 62.56 & 50.26 &
        -- & -- & -- & -- & -- & 74.87 \\
        Qwen2.5-3B-EntropyLong-128K & -- & 128K & 62.02 & 49.11 &
        -- & -- & -- & -- & -- & 74.93 \\
        Qwen2.5-3B-PolicyLong-128K & -- & 128K & 64.10 & 51.53 &
        -- & -- & -- & -- & -- & 76.67 \\
        \hline
        \multicolumn{11}{l}{\textit{RSI}} \\
        Llama3-8B & \multirow{2}{*}{RRSI} & 8K & 39.67 & 46.02 &
        28.19 & \textbf{64.63} & 83.45 & 22.18 & 31.65 & 33.32 \\
        Qwen2.5-3B & & 32K & 38.46 & 44.83 &
        39.11 & 56.88 & 79.47 & 21.15 & 27.55 & 32.09 \\
        \hline
        \multicolumn{11}{l}{\textit{Ours}} \\
        Llama3-8B & \multirow{2}{*}{LongCapHarness} & 8K & \cellcolor{topcolor}\textbf{79.32} & \textbf{64.01} &
        \textbf{100.00} & 59.70 & \textbf{83.85} & 39.67 & 36.84 & \textbf{94.62} \\
        Qwen2.5-3B & & 32K & 78.73 & 63.18 &
        \textbf{100.00} & 56.33 & 76.84 & \textbf{56.28} & 26.45 & 94.28 \\
        \hline
    \end{tabular}}
\end{table*}

We aim to answer four main questions: \emph{Q1:} How well does \emph{LongCapHarness} perform, and what explains its strengths and limitations? \emph{Q2:} Does \emph{LongCapHarness} generalize across different LLM execution models? \emph{Q3:} How does capability primitive RSI affect task performance? \emph{Q4:} How cost-effective is \emph{LongCapHarness}?

\subsection{Experimental Setup}
\label{sec:experimental-setting}

\paragraph{Benchmarks.}
We use HELMET \cite{yen2025helmet}, RULER
\cite{hsieh2024ruler} and LongBench-v2~\citep{bai2025longbenchv2} to evaluate \emph{LongCapHarness} under different context lengths.

We construct the development set based on the principle of
exposing harness development to each evaluated task type while
testing reuse on unseen inputs. The resulting split contains
2,242 development examples and 59,826 held-out test examples,
with no overlap between the two sets. Appendix~\ref{sec:data-splits} details the development set and test set for each benchmarks.

\paragraph{Models and baselines.}
We select Llama3-8B~\cite{meta2024llama3} and Qwen2.5-3B~\cite{yang2024qwen25} as execution models for \emph{LongCapHarness} and RRSI~\cite{xia2026rrsi}. We compare \emph{LongCapHarness} and RRSI results against the base models and their respective 128K context-extension variants from NExtLong~\cite{gao2025nextlong} and PolicyLong~\cite{jia2026policylong}.
We also include EntropyLong, using the Llama results from EntropyLong paper~\cite{jia2026entropylong} and the Qwen results from the PolicyLong paper~\cite{jia2026policylong}, averaged across 8K--128K. 
During development, both \emph{LongCapHarness} and RRSI use GPT-5.6-Sol as the developer model and Llama3-8B for candidate execution. In RRSI, GPT-5.6-Sol serves as the proposer, selector, critic, and pruner.

\paragraph{Harnessed primitives.}
LongCapHarness for HELMET and RULER uses keyword matching, lookup,
relation following, state tracking, counting, ordering, relevance
judgement, rule inference, classification, evidence selection, and
evidence integration. Appendix~\ref{sec:harness-capabilities} details the capability
decompositions, task mappings, and each primitive's specification, inputs, and outputs.

\paragraph{Metrics and aggregation.}
We report task scores on a 0--100 scale~(the accuracy rate~$\uparrow$). Metrics include substring exact match for RAG, exact match for ICL and InfiniteBench Choice, NDCG@10 for reranking, and the designated recall or QA metric for RULER. InfiniteBench QA and NarrativeQA use ROUGE-L F1. 
Category scores average each task across input lengths, then average the tasks within the category.

\subsection{Main Results}
\label{sec:main-results}

\subsubsection{Comparison of the Three Approaches}
\label{sec:retained-results}

Table~\ref{tab:main-results} compares \emph{LongCapHarness} with base
models, model-training methods, and RRSI. \emph{LongCapHarness}
achieves the highest aggregate score among the evaluated
methods for both execution models. With Llama3-8B,
\emph{LongCapHarness} outperforms the base model, NExtLong, and
RRSI by 64.18, 9.17, and 39.65 points.
With Qwen2.5-3B, it outperforms the base model, PolicyLong,
and RRSI by 16.82, 14.63, and 40.27 points.
These results demonstrate the effectiveness of \emph{LongCapHarness}
on the evaluated long-context benchmarks.

RRSI, however, falls below NExtLong by 30.48 points with
Llama and below PolicyLong by 25.64 points with Qwen.
Inspection of RRSI's intermediate development code shows
repeated attempts to address context overflow through token
limits and character-based chunking. Within the available
development rounds, these revisions approach task solving
primarily through the physical organization of the input and organize operators around it. In contrast, LongCapHarness starts from the logical
organization of task-solving operations. \textbf{Q1 Strength: }The capability-
decomposition requirements make these operations and their
dependencies explicit, guiding RSI to optimize over the task's computational structure.

\begin{figure}[!t]
    \centering
    \includegraphics[width=0.95\linewidth]{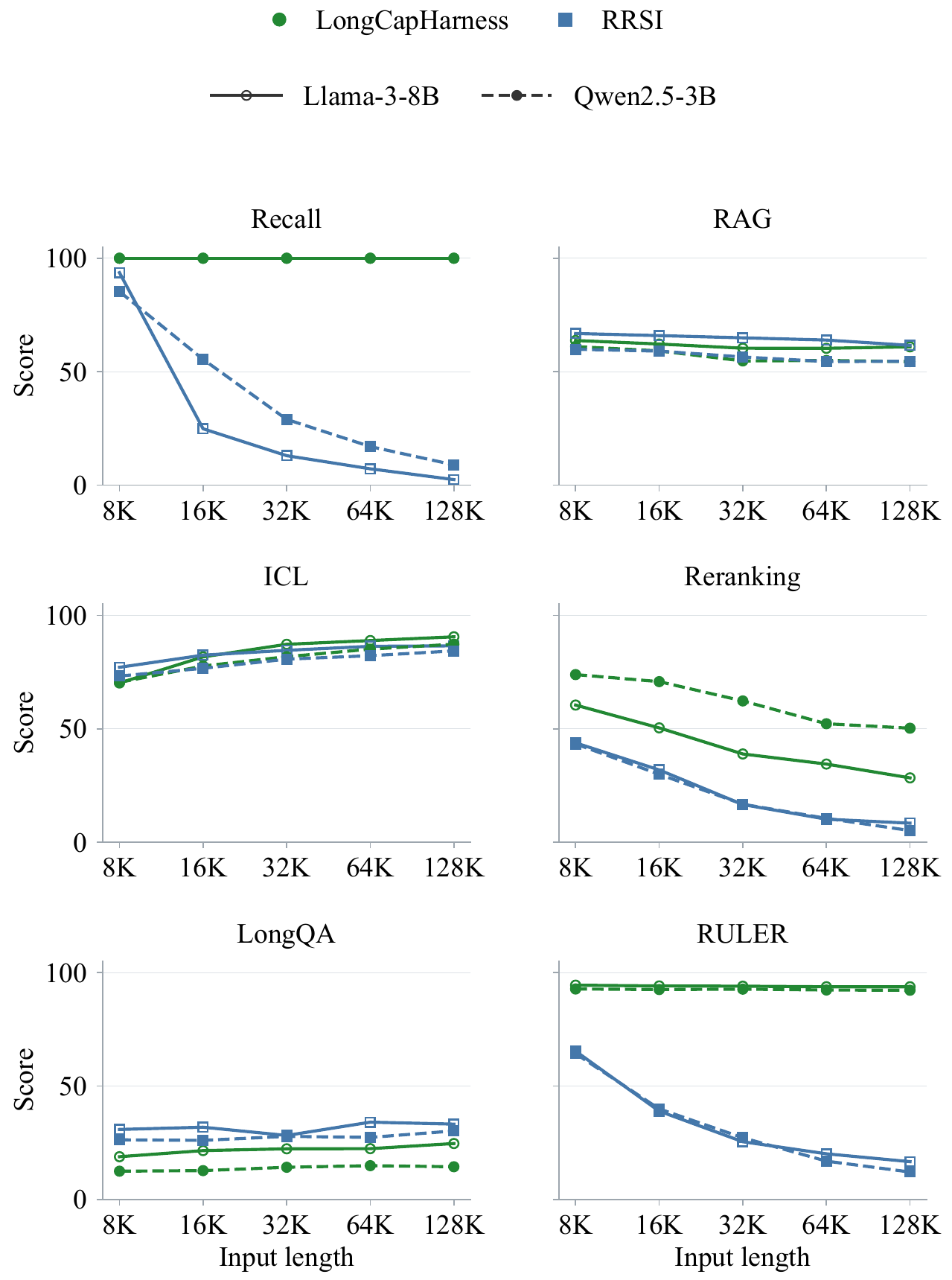}
    \vspace{-5pt}
    \caption{Helmet and Ruler category scores across input lengths with Llama3-8B and Qwen2.5-3B.}
    \label{fig:length-results}
    \vspace{-5pt}
\end{figure}

We next examine the
task categories to understand where this approach yields
the gains and where its benefits remain limited.

The largest gain appears in Recall,
where LongCapHarness achieves 100.00 with both execution
models, outperforming RRSI by 71.81 points with Llama
and 60.89 points with Qwen, and NExtLong by 17.44 points.
Figure~\ref{fig:length-results} shows that this performance
persists across the evaluated 8K--128K input lengths.
The advantage extends to RULER, where LongCapHarness
achieves 94.62 with Llama and 94.28 with Qwen,
outperforming RRSI by 61.30 and 62.19 points and the
corresponding context-extension models by 13.12 and
17.61 points.
\textbf{Q1 Strength: }These gains stem from the fact that the underlying primitives have explicit rules that code can execute across longer inputs, without requiring the model to maintain the complete task state in its context. For example, Recall uses keyword matching and lookup, while most RULER tasks use relation following, state tracking, and counting.

The improvements are not limited to tasks that code can
solve directly. On reranking, LongCapHarness outperforms
RRSI by 17.49 points with Llama and 35.13 points with
Qwen; Llama also exceeds NExtLong by 8.20 points.
\textbf{Q1 Strength: }The harness combines model-based relevance judgement with code-controlled ordering, leaving semantic judgements as a limiting factor. Section~\ref{sec:capability-primitive-evolution} examines the effects of different primitive implementations.

The performance gap is smaller and less consistent on
ICL, RAG, and LongQA. On ICL, LongCapHarness exceeds
RRSI by 0.40 points with Llama and falls below it by
2.63 points with Qwen. On RAG, it trails RRSI by
4.93 and 0.55 points, respectively. On LongQA, Llama
improves over RRSI by 5.19 points but remains 0.46 points
below NExtLong, while Qwen falls below RRSI by 1.10 points. 
These mixed gains are consistent with continued reliance on model capabilities: inferring mappings from demonstrations in ICL, and identifying and integrating evidence in RAG and LongQA. \textbf{Q1 Limitation: } Harnesses organize these operations but do not replace their semantic judgements.

\paragraph{LongBench-v2.}
We report our Qwen2.5-3B results on the Medium and Long subsets of
LongBench-v2~\citep{bai2025longbenchv2} in Table~\ref{tab:longbench-v2},
with input lengths of 32K--128K and 128K--2M tokens.
LongCapHarness scores 34.47 on Medium and 27.85 on Long,
exceeding PolicyLong by 7.37 points on Medium but trailing it
by 3.65 points on Long. 
These results are consistent with the limitation discussed above.

\begin{table}[!t]
    \centering
    \footnotesize
    \caption{\textbf{LongBench-v2 score with Qwen2.5-3B.}
    Medium and Long denote input lengths of 32K--128K and 128K--2M tokens.
    Published baseline scores are post-training results from
    \cite{jia2026policylong}.}
    \label{tab:longbench-v2}
    \begin{tabular*}{\linewidth}{@{\extracolsep{\fill}}lrr@{}}
        \toprule
        Method & Medium & Long \\
        \midrule
        Base & 21.50 & 20.60 \\
        NExtLong & 23.70 & 26.90 \\
        EntropyLong & 24.20 & 28.70 \\
        PolicyLong & 27.10 & \textbf{31.50} \\
        LongCapHarness & \textbf{34.47} & 27.85 \\
        \bottomrule
    \end{tabular*}
\end{table}

\subsubsection{Cross-Model Transfer}
\label{sec:cross-model-transfer}

Replacing Llama3-8B with Qwen2.5-3B decreases
LongCapHarness's Avg by only 0.59 points, from 79.32
to 78.73. The developed harnesses are reused with
adaptations to the model, tokenizer, evidence budget,
and serving configuration.

Code-based primitives follow the same execution rules
across models, as illustrated by Recall remaining at
100.00. Semantic primitives rely on the new model's
judgements, leading to task-level differences: reranking
improves by 16.61 points with Qwen, while LongQA
decreases by 10.39 points.
\textbf{Q2 Answer:} Cross-model generalization is strong for the code-based
components of capability harnesses, while semantic
components depend on the execution model's capabilities.

\subsubsection{Capability Primitive Evolution}
\label{sec:capability-primitive-evolution}

We examine the RSI trajectory of capability primitives
by HELMET Reranking. The target capability remains
fixed, while its relevance-judgement and ordering primitives
evolve through three implementations:
BM25 scoring with numeric sorting~(V0),
independent LLM scoring with numeric sorting~(V1),
and relative LLM comparison with group ordering and
merging~(V2). Table~\ref{tab:reranking-evolution} reports
their performance across execution models and input lengths.

The four LLM-based runs are evaluated on the same
500 records, with 100 at each input length from 8K
to 128K. Each implementation is fixed before evaluation
and processes the complete candidate set within the
execution model's per-call context budget.

\begin{table}[!t]
    \centering
    \footnotesize
    \setlength{\tabcolsep}{2pt}
    \caption{\textbf{RSI trajectory of HELMET reranking
    capability primitives.}
    Bold indicates the highest displayed score in each column.}
    \label{tab:reranking-evolution}
    \begin{tabularx}{\linewidth}{@{}>{\raggedright\arraybackslash}Xrrrrrr@{}}
        \toprule
        Version & 8K & 16K & 32K & 64K & 128K & Mean \\
        \midrule
        \multicolumn{7}{l}{Initial code-based implementation (both models)} \\
        V0: BM25 + sort & 53.26 & 45.44 & 39.52 & 34.66 & 28.85 & 40.35 \\
        \midrule
        \multicolumn{7}{l}{Llama3-8B} \\
        V1: LLM scoring + sort & 65.51 & 51.88 & 36.50 & 22.95 & 17.89 & 38.95 \\
        V2: Comparison + merge & 57.79 & 47.48 & 36.70 & 30.85 & 25.51 & 39.67 \\
        \midrule
        \multicolumn{7}{l}{Qwen2.5-3B} \\
        V1: LLM scoring + sort & 68.14 & 55.13 & 37.06 & 25.12 & 17.08 & 40.50 \\
        V2: Comparison + merge & \textbf{71.76} & \textbf{64.48} & \textbf{55.50} & \textbf{45.85} & \textbf{43.83} & \textbf{56.28} \\
        \bottomrule
    \end{tabularx}
\end{table}

The first evolution, from V0 to V1, changes the
relevance-judgement primitive from lexical matching to
semantic scoring. The LLM assigns each query--passage
pair a relevance grade, while the ordering primitive
continues to sort the resulting scores in code.
This change improves performance at 8K--16K for both
models, while V0 retains higher scores at 32K--128K.
Across all lengths, V1 achieves 38.95 with Llama and
40.50 with Qwen, compared with V0's 40.35.
This step introduces semantic interpretation into the
relevance primitive, with benefits concentrated at
shorter input lengths.

The second evolution, from V1 to V2, changes both the
relevance output and its downstream composition.
The LLM expresses relative preferences among passages,
and the ordering primitive uses these preferences to
order small groups and merge them into a complete ranking.
With Qwen, this evolution improves the mean by 15.78
points to 56.28 and achieves the highest score at every
input length. At 128K, the gain over V1 reaches 26.75
points. With Llama, the mean improves by 0.72 points
to 39.67, with gains at 32K--128K; V1 retains higher
scores at 8K--16K.

\textbf{Q3 Answer: }These results show that capability primitive RSI improves task performance by
jointly refining local judgements and their composition,
with gains depending on the execution model and input length.

\subsubsection{Cost Effectiveness}
\label{sec:cost-efficiency}

We report development and inference usage for \emph{LongCapHarness} and RRSI, alongside model-training rental estimates.
Development costs are estimated token charges for GPT-5.6 Sol, and
training costs are estimated H100 rental charges. Llama and Qwen inference
usage is reported as input and output tokens, and token charges
estimated at reference API rates.
Development uses GPT-5.6 Sol API rates, inference uses reference API rates, and training uses H100 rental rates. These are estimated charges; pricing sources and assumptions are given in Appendix~\ref{sec:cost-estimation-details}.


\paragraph{Development costs.}
Table~\ref{tab:token-costs} shows that LongCapHarness uses 814.11M input
and 3.35M output tokens during development, compared with 52.42M and
3.87M for RRSI, yielding estimated API charges of \$541.47 and \$331.56.

NExtLong's reported training configuration costs an estimated \$3,696.00. PolicyLong's FLOP-scaled estimate is \$1,858.82, assuming equal effective throughput per GPU.
Overall, LongCapHarness has higher development API charges than RRSI but lower estimated costs than either model-training reference.

\begin{table}[!t]
    \centering
    \footnotesize
    \setlength{\tabcolsep}{2pt}
    \caption{\textbf{Development, training, and inference costs.}
    Input and output are in millions of tokens; costs are estimated USD.
    Training rows show H100 GPU-hours and rental costs; PolicyLong uses
    the FLOP-scaling estimate. Bold indicates lower token usage or cost
    within each group; pink highlights LongCapHarness's better values.}
    \label{tab:token-costs}
    \begin{tabularx}{\linewidth}{@{}>{\raggedright\arraybackslash}Xrrr@{}}
        \toprule
        Method & \multicolumn{2}{c}{H100 GPU-hours} & \shortstack{Cost\\(\$)} \\
        \midrule
        \multicolumn{4}{l}{\textit{Model Training}} \\
        NExtLong~\citep{gao2025nextlong} & \multicolumn{2}{c}{960} & 3,696.00 \\
        PolicyLong~\citep{jia2026policylong} & \multicolumn{2}{c}{482.81} & \textbf{1,858.82} \\
        \midrule[1pt]
        Method & \shortstack{Input\\(M)} & \shortstack{Output\\(M)} & \shortstack{Cost\\(\$)} \\
        \midrule
        \multicolumn{4}{l}{\textit{Development (GPT-5.6 Sol)}} \\
        RRSI~\citep{xia2026rrsi} & \textbf{52.42} & 3.87 & \textbf{331.56} \\
        \emph{LongCapHarness} & 814.11 & \cellcolor{topcolor}\textbf{3.35} & 541.47 \\
        \midrule
        \multicolumn{4}{l}{\textit{Inference (Llama-3-8B)}} \\
        RRSI~\citep{xia2026rrsi} & \textbf{236.05} & 89.55 & 8.30 \\
        \emph{LongCapHarness} & 408.00 & \cellcolor{topcolor}\textbf{1.87} & \cellcolor{topcolor}\textbf{8.23} \\
        \midrule
        \multicolumn{4}{l}{\textit{Inference (Qwen2.5-3B)}} \\
        RRSI~\citep{xia2026rrsi} & \textbf{285.61} & 8.33 & \textbf{29.39} \\
        \emph{LongCapHarness} & 535.08 & \cellcolor{topcolor}\textbf{2.34} & 53.74 \\
        \bottomrule
    \end{tabularx}
\end{table}

\paragraph{Inference costs.}

We compare LongCapHarness and RRSI on the same
benchmark test set.
Table~\ref{tab:token-costs} shows that LongCapHarness uses more input
tokens but fewer output tokens with both models. At the reference
API rates, its estimated charges are slightly lower with Llama
(\$8.23 vs.\ \$8.30) but higher with Qwen (\$53.74 vs.\ \$29.39).

\textbf{Q4 Answer: }In conclusion, LongCapHarness invests more development
tokens in finding, testing, and freezing reusable code
implementations in the capability harness. During inference,
these implementations execute operations that would otherwise
require model reasoning, reducing generated output tokens.

%% file: secs/sec5.1_limit.tex
\section{Conclusion}

\emph{LongCapHarness} decomposes long-context tasks into capability primitives and uses RSI to build reusable harnesses without data synthesis or model post-training. Transferred from Llama3-8B to Qwen2.5-3B, it improves average HELMET and RULER scores over model-training baselines; its estimated development cost is below the estimated training costs for NExtLong and PolicyLong. These results support harnessing latent LLM capabilities beyond explicit training.

\section{Limitation} 

\emph{LongCapHarness} still relies on a powerful harnessing agent (GPT-5.6-Sol) to construct its capability harnesses. We observe a sharp performance drop when weaker models such as Llama3-8B and Qwen2.5-3B serve as their own harnessing agents. Enabling weaker models to develop effective harnesses independently remains an open challenge.

\section{Ethical Considerations}

Our method constructs and refines capability harnesses for
long-context reasoning without additional LLM training.
All benchmark datasets used in this study are publicly
available and contain no personal information. No human
subjects are involved, and all LLMs used are publicly
available models or accessible through public APIs.
We do not identify additional ethical concerns specific
to the proposed method within this experimental setting.

%% file: secs/sec6_appendix.tex
\section{Harnessed Capabilities and Primitives}
\label{sec:harness-capabilities}

The following list summarizes the capability decompositions
described for our HELMET and RULER harnesses.
Capabilities are grouped by the transformation required to solve a task. 
Primitive names describe transformations, not particular code or prompt
implementations. Multiple task families can share a decomposition.

\begin{itemize}
	\item \textbf{Matching and retrieval.} \emph{Keyword matching} and \emph{lookup} of
	matching records or values support HELMET Recall and RULER needle retrieval.
	\item \textbf{Relation and state tracking.} \emph{Relation following} across
	records and \emph{state tracking} through updates support RULER variable tracking.
	\item \textbf{Frequency-based selection.} \emph{Counting} occurrences and
	\emph{ordering} by frequency support RULER common/frequent word extraction.
	\item \textbf{Passage reranking.} \emph{Relevance judgement} and \emph{ordering}
	candidates support HELMET Reranking.
	\item \textbf{Example-based classification.} \emph{Rule inference} from
	examples and \emph{classification} into allowed labels support HELMET ICL.
	\item \textbf{Evidence-based answering.} \emph{Evidence selection} of relevant
	passages and \emph{evidence integration} into an answer support HELMET RAG and LongQA,
	and RULER QA.
\end{itemize}

\subsection{Primitive Specifications and Interfaces}

For each primitive, \emph{Spec} states the intended transformation,
while \emph{Input} and \emph{Output} define its interface. These are
abstract requirements, not implementation-specific data formats or
guarantees of model accuracy.

\begin{itemize}
	\item \textbf{Keyword matching.}
	\emph{Spec:} Locate occurrences satisfying the supplied lexical matching rules.
	\emph{Input:} Keywords, text records, and matching rules.
	\emph{Output:} Matching spans and their record identities.

	\item \textbf{Lookup.}
	\emph{Spec:} Retrieve records or values satisfying a search condition.
	\emph{Input:} Search condition and records.
	\emph{Output:} Matching records or values.

	\item \textbf{Relation following.}
	\emph{Spec:} Follow the specified relations from a starting entity
	across records to recover the requested entity or value.
	\emph{Input:} Starting entity, relations to follow, and records.
	\emph{Output:} Related entity or value.

	\item \textbf{State tracking.}
	\emph{Spec:} Apply events according to the supplied update rules
	and event order to determine the state at a requested time.
	\emph{Input:} Initial state, events, update rules, and requested time.
	\emph{Output:} State at the requested time.

	\item \textbf{Counting.}
	\emph{Spec:} Aggregate occurrences according to the supplied counting rules.
	\emph{Input:} Records and counting rules.
	\emph{Output:} A count or per-item counts.

	\item \textbf{Ordering.}
	\emph{Spec:} Arrange candidates according to the supplied scores,
	preferences, or ordering rules while preserving candidate identities.
	\emph{Input:} Candidates and ordering rules, including any required
	scores or comparison outputs.
	\emph{Output:} Ordered candidate IDs.

	\item \textbf{Relevance judgement.}
	\emph{Spec:} Assess how well a passage addresses the query, or
	compare the relevance of candidate passages.
	\emph{Input:} Query and passage(s).
	\emph{Output:} Relevance score or passage preference.

	\item \textbf{Rule inference.}
	\emph{Spec:} Infer a mapping compatible with the supplied
	input--output examples and any candidate-rule constraints.
	\emph{Input:} Input--output examples and candidate rules, when supplied.
	\emph{Output:} Inferred mapping or compatible rule(s).

	\item \textbf{Classification.}
	\emph{Spec:} Assign an input to an allowed label using the supplied
	examples or inferred mapping.
	\emph{Input:} Input to classify, allowed labels, and examples or mapping.
	\emph{Output:} Predicted label.

	\item \textbf{Evidence selection.}
	\emph{Spec:} Identify passages or spans relevant to answering the
	question while retaining their source identities.
	\emph{Input:} Question, candidate passages, and source IDs.
	\emph{Output:} Selected evidence with source IDs.

	\item \textbf{Evidence integration.}
	\emph{Spec:} Combine information from the supplied evidence to
	answer the question, preserving the relationships needed for the answer.
	\emph{Input:} Question, selected passages or evidence, and source IDs.
	\emph{Output:} Answer.
\end{itemize}

\section{Cost Estimation Details}
\label{sec:cost-estimation-details}

\paragraph{API-price estimates.}
For development, we apply GPT-5.6 Sol Standard short-context rates of
\$4/\$0.40/\$20 per million ordinary-input/cache-read/output
tokens for requests with at most 272K input tokens
\citep{openai2026solpricing}. For inference, we price Llama3-8B at
\$0.02/\$0.04 per million input/output tokens using DeepInfra's
Llama 3.1 8B Instruct quotation on OpenRouter
\citep{openrouter2026llamapricing}, and Qwen2.5-3B at \$0.10/\$0.10
using the Qwen2.5 Coder 3B quotation on OpenRouter
\citep{openrouter2026qwenpricing}.

\paragraph{Training rental estimates.}
At an H100 rental rate of \$3.85 per GPU-hour
\citep{nebius2026pricing}, NExtLong's reported 64 GPUs
for 15 hours \citep{gao2025nextlong} cost an estimated
\$3,696.00. Scaling by approximate training FLOPs $F\approx 6T(P+Lsd)$,
where $T$ is the token budget, $P$ the parameter count, $L$ the layer
count, $s$ the sequence length, and $d$ the hidden width, and assuming
equal effective throughput per GPU gives a PolicyLong estimate of
\$1,858.82.

\section{Development and Evaluation Data}
\label{sec:data-splits}

Table~\ref{tab:split-overview} breaks down the 2,242 development and
59,826 test/evaluation records reported in
Section~\ref{sec:experimental-setting} by benchmark. The following
tables provide category, subtask, and input-length breakdowns.

\begin{table}[!t]
	\centering
	\footnotesize
	\caption{Development and test/evaluation records by benchmark.}
	\label{tab:split-overview}
	\begin{tabular*}{\linewidth}{@{\extracolsep{\fill}}lrrr@{}}
		\toprule
		Benchmark & Development & Test/evaluation & Total \\
		\midrule
		HELMET & 2,080 & 27,035 & 29,115 \\
		RULER & 112 & 32,388 & 32,500 \\
		LongBench-v2 & 50 & 403 & 453 \\
		\midrule
		Total & 2,242 & 59,826 & 62,068 \\
		\bottomrule
	\end{tabular*}
\end{table}

\subsection{HELMET and RULER}

\begin{table}[!t]
	\centering
	\footnotesize
	\caption{Full HELMET/RULER split by category, aggregated over five input lengths.}
	\label{tab:split-categories}
	\begin{tabular*}{\linewidth}{@{\extracolsep{\fill}}lrrr@{}}
		\toprule
		Category & Development & Test & Total \\
		\midrule
		Recall & 70 & 1,930 & 2,000 \\
		RAG & 945 & 9,555 & 10,500 \\
		ICL & 75 & 12,425 & 12,500 \\
		Re-rank & 210 & 405 & 615 \\
		LongQA & 450 & 1,050 & 1,500 \\
		Citation & 30 & 970 & 1,000 \\
		Summarization & 300 & 700 & 1,000 \\
		\midrule
		HELMET subtotal & 2,080 & 27,035 & 29,115 \\
		RULER & 112 & 32,388 & 32,500 \\
		\midrule
		Total & 2,192 & 59,423 & 61,615 \\
		\bottomrule
	\end{tabular*}
\end{table}

Both partitions cover 34 subtasks at five benchmark input lengths,
or 170 task--length cells each. These lengths describe benchmark
inputs, not the context budget of an individual LLM call.
Tables~\ref{tab:split-lengths} and~\ref{tab:split-subtasks} report the
length and subtask breakdowns; the latter sums over all five lengths.

\begin{table}[!t]
	\centering
	\footnotesize
	\caption{Full HELMET/RULER split by benchmark input length.}
	\label{tab:split-lengths}
	\begin{tabular*}{\linewidth}{@{\extracolsep{\fill}}lrrr@{}}
		\toprule
		Input length & Development & Test & Total \\
		\midrule
		8K (8,192) & 458 & 11,865 & 12,323 \\
		16K (16,384) & 431 & 11,892 & 12,323 \\
		32K (32,768) & 438 & 11,885 & 12,323 \\
		64K (65,536) & 433 & 11,890 & 12,323 \\
		128K (131,072) & 432 & 11,891 & 12,323 \\
		\midrule
		Total & 2,192 & 59,423 & 61,615 \\
		\bottomrule
	\end{tabular*}
\end{table}

\begin{table}[!t]
	\centering
	\footnotesize
	\setlength{\tabcolsep}{3pt}
	\caption{All 34 HELMET/RULER subtasks in the full split.}
	\label{tab:split-subtasks}
	\begin{tabularx}{\linewidth}{@{}l>{\raggedright\arraybackslash}Xrrr@{}}
		\toprule
		Category & Subtask & Dev. & Test & Total \\
		\midrule
		Recall & json\_kv & 54 & 446 & 500 \\
		Recall & niah\_multikey\_2 & 5 & 495 & 500 \\
		Recall & niah\_multikey\_3 & 5 & 495 & 500 \\
		Recall & niah\_multivalue & 6 & 494 & 500 \\
		RAG & rag\_hotpotqa & 45 & 1,455 & 1,500 \\
		RAG & rag\_natural\_questions & 90 & 2,910 & 3,000 \\
		RAG & rag\_popqa & 720 & 2,280 & 3,000 \\
		RAG & rag\_triviaqa & 90 & 2,910 & 3,000 \\
		ICL & icl\_banking77 & 15 & 2,485 & 2,500 \\
		ICL & icl\_clinic150 & 15 & 2,485 & 2,500 \\
		ICL & icl\_nlu & 15 & 2,485 & 2,500 \\
		ICL & icl\_trec\_coarse & 15 & 2,485 & 2,500 \\
		ICL & icl\_trec\_fine & 15 & 2,485 & 2,500 \\
		Re-rank & rerank\_msmarco & 210 & 405 & 615 \\
		LongQA & infbench\_choice & 155 & 345 & 500 \\
		LongQA & infbench\_qa & 145 & 355 & 500 \\
		LongQA & narrativeqa & 150 & 350 & 500 \\
		Citation & alce\_asqa & 15 & 485 & 500 \\
		Citation & alce\_qampari & 15 & 485 & 500 \\
		Summarization & infbench\_sum\_eng & 150 & 350 & 500 \\
		Summarization & multi\_lexsum & 150 & 350 & 500 \\
		RULER & ruler\_cwe & 5 & 2,495 & 2,500 \\
		RULER & ruler\_fwe & 5 & 2,495 & 2,500 \\
		RULER & ruler\_niah\_multikey\_1 & 8 & 2,492 & 2,500 \\
		RULER & ruler\_niah\_multikey\_2 & 5 & 2,495 & 2,500 \\
		RULER & ruler\_niah\_multikey\_3 & 5 & 2,495 & 2,500 \\
		RULER & ruler\_niah\_multiquery & 7 & 2,493 & 2,500 \\
		RULER & ruler\_niah\_multivalue & 7 & 2,493 & 2,500 \\
		RULER & ruler\_niah\_single\_1 & 13 & 2,487 & 2,500 \\
		RULER & ruler\_niah\_single\_2 & 14 & 2,486 & 2,500 \\
		RULER & ruler\_niah\_single\_3 & 7 & 2,493 & 2,500 \\
		RULER & ruler\_qa\_1 & 15 & 2,485 & 2,500 \\
		RULER & ruler\_qa\_2 & 15 & 2,485 & 2,500 \\
		RULER & ruler\_vt & 6 & 2,494 & 2,500 \\
		\midrule
		Total & 34 subtasks & 2,192 & 59,423 & 61,615 \\
		\bottomrule
	\end{tabularx}
\end{table}

\subsection{LongBench-v2}

Table~\ref{tab:longbench-domains} breaks down the 50 development
and 403 held-out evaluation records across five non-code domains,
with ten development records per domain.

\begin{table}[t]
	\centering
	\footnotesize
	\caption{LongBench-v2 split by domain. Test records exclude the development subset.}
	\label{tab:longbench-domains}
		\begin{tabularx}{\linewidth}{@{}>{\raggedright\arraybackslash}Xrrr@{}}
		\toprule
			Domain & Development & Test & Total \\
		\midrule
			Single-Document QA & 10 & 165 & 175 \\
			Multi-Document QA & 10 & 115 & 125 \\
			Long In-context Learning & 10 & 71 & 81 \\
			Long-dialogue History Understanding & 10 & 29 & 39 \\
			Long Structured Data Understanding & 10 & 23 & 33 \\
		\midrule
			Total & 50 & 403 & 453 \\
		\bottomrule
	\end{tabularx}
\end{table}